\def\year{2022}\relax
\documentclass[letterpaper]{article} 
\usepackage{aaai22}  
\usepackage{times}  
\usepackage{helvet}  
\usepackage{courier}  
\usepackage[hyphens]{url}  
\usepackage{graphicx} 
\usepackage{natbib}  
\usepackage{caption} 
\DeclareCaptionStyle{ruled}{labelfont=normalfont,labelsep=colon,strut=off} 
\usepackage{algorithm}
\usepackage{algorithmic}
\usepackage{adjustbox}
\usepackage{amssymb}

\usepackage{newfloat}
\usepackage{listings}
\floatstyle{ruled}
\newfloat{listing}{tb}{lst}{}
\floatname{listing}{Listing}
\title{GKS: Graph-based Knowledge Selector for Task-oriented Dialog System}
\author{
    Jen-Chieh Yang\equalcontrib\textsuperscript{\rm 1},
    Jia-Yan Wu\equalcontrib\textsuperscript{\rm 2},
    Sung-Ping Chang\textsuperscript{\rm 3},
    Ya-Chieh Huang\textsuperscript{\rm 2}
}
\affiliations{
    \textsuperscript{\rm 1} Academic Sinica\\
    \textsuperscript{\rm 2} National Taiwan University\\
    \textsuperscript{\rm 3} Columbia University\\
    richardyangms3@gmail.com, b06701216@ntu.edu.tw,  sksk29292929@gmail.com,
    b06701235@ntu.edu.tw
}

\usepackage{bibentry}

\begin{document}

\maketitle

\begin{abstract}
In previous research, knowledge-selection tasks mostly rely on language model-based methods or knowledge ranking. However, while approaches that rely on the language models take all knowledge as sequential input, knowledge does not contain sequential information in most circumstances. On the other hand, the knowledge-ranking methods leverage dialog history and each given knowledge snippet separately, but they do not consider information between knowledge snippets. In the Tenth Dialog System Technology Challenges (DSTC10), we participated in the second Knowledge-grounded Task-oriented Dialogue Modeling on Spoken Conversations. To deal with the problems mentioned above, we modified training methods based on state-of-the-art (SOTA) models for the first and third sub-tasks. As for the second sub-task of knowledge selection, we proposed Graph-Knowledge Selector (GKS), utilizing a graph-attention base model incorporated with the language model. GKS makes knowledge-selection decisions in the dialog by simultaneously considering each knowledge embedding generated from the language model without sequential features. Moreover, GKS leverages considerable knowledge in decision-making and takes relations across knowledge as part of the selection process. As a result, GKS outperforms several SOTA models proposed in the data-set on knowledge selection from the Ninth Dialog System Technology Challenges (DSTC9).
\end{abstract}

\section{Introduction}
Nowadays, task-oriented dialog systems have been widely used, providing specific services in different industries. Therefore, building dialog systems to deal with heterogeneous data becomes significant to make robust and general frameworks for wide-range applied scenarios. The problem amplifies when input data is considered to be spoken language, making input data with even more variance. In this paper, we proposed our approach to constructing a dialog system in each sub-task to alleviate differentiation for data sets. We mainly focus on the Knowledge-selection task, aiming to solve potential obstacles past knowledge selection might encounter.

In DSTC10 track 2, the goal is to perform a dialog system on spoken language, which is not straightforward via a pre-trained language model. Regarding task 1, we design our knowledge-seeking turn detection model with a denoise language model. The model is pre-trained with text data, aiming to transfer spoken language to text-style data.

Knowledge selection approach in the past mainly utilized language models. These approaches often have minor or no change on model structure but training objective and some training techniques, such as data augmentation. Here, we assume information within knowledge for dialog should not be taken simply in a sequential style language model (see. Table \ref{example}).To better capture knowledge information, we form our knowledge selection model with a graph-based design for task 2. The proposed Graph-Knowledge Selector (GKS) takes knowledge-embedding as input to make selection tasks. GKS does not simply concatenate knowledge and question together into a language model as input and output prediction. Instead, GKS leverages information between each knowledge and question with a graph-attention model.

To assure our proposed frameworks in previous tasks, we select a model proposed in DSTC9 without much modification. Therefore, we choose \cite{bao2020plato} with modification as our generation model, which takes the output from knowledge selection task prediction as input.

The main contribution in this paper is a new knowledge selection model for the dialog system. However, since our approach does not solve the problem of spoken language quite well, we further test our framework on DSTC9 data set, which consists of text data. The experiment result shows that our proposed framework outperforms models proposed in last year's challenge. Our implementation will be released upon acceptance.

\begin{table}[ht]
\centering
\begin{adjustbox}{width={\columnwidth},center}%
\begin{tabular}{ll}
\hline {}
Detected turn&\textbf{Does the hotel offer accessible parking?} \\ 
\hline
Knowledge0 & Does the hotel offer accessible parking?\\
Knowledge1 & Is there on-site private parking at the
\\ &  Bridge Guest House?\\
Knowledge2 & Do I have to pay for parking?\\ 
Knowledge3 & Is there a cost for parking?\\ 
Knowledge4 & Can I make a reservation for parking?\\
Knowledge5 & Do they have help for disabled parking?\\
Knowledge6 & Do you provide room service daily? \\
Knowledge7 & Are there any fitness center or gym?\\
\hline
\end{tabular}
\end{adjustbox}
\caption{Example from dialog turn and knowledge. The desired knowledge for the user should be Knowledge0. However, there are no sequential relations between knowledge, but many approaches simply connect knowledge as input. Moreover, if we train the model simply pairing Detected Turn with each knowledge and sampling training pairs, the overlapping information between knowledge might be wasted. In the example, Knowledge0-5 refers to parking issues; we try to let the model distinguish Knowledge0 from all the others according to detected turn but not with sampling approaches.}
\label{example}
\end{table}

\section{Related Work}
The recent development of task-oriented dialogue systems benefits from pre-trained language models like GPT-2 ~\citep{Radford:18, Budzianowski:19, Ham:20}. Furthermore, BERT-like ~\citep{Devlin:19} architectures, such as RoBERTa ~\citep{Yinhan:19}, achieved state-or-the-art performance on natural language understanding tasks like GLUE dataset ~\citep{Wang:18}. In subtasks 1 and 3, we leveraged this architecture to attain better performance.

\subsection{Knowledge-Seeking Turn Detection}
The knowledge-seeking turn detection was first introduced in DSTC9 track1 by ~\citep{kim2020domain}. A binary classifier was proposed to solve this task, while ~\citet{Tang:21} used a supervised method with an ELECTRA-based model followed by a fully connected layer ~\citep{clark2020electra} to determine whether to trigger knowledge. In another way, ~\citet{Mi:21} employed an ensemble algorithm of four pre-trained models, such as RoBERTa and UniLM ~\citep{dong2019unified}, to solve it as a classification task as well. ~\citet{bao2020plato} proposed to consider API and external knowledge by schema description method. ~\citep{shah2019robust, eric2019multiwoz, rastogi2020schemaguided}

\subsection{Knowledge Selection}
The knowledge selection task is to retrieve candidate snippets from the knowledge base for response generation. Traditionally, TF-IDF technique~\citep{Ramos2003UsingTT} and language model are applied on similar tasks. As mentioned above, the limitations of previous works lie in the model structure. On the other hand, we are inspired by Kernel Graph Attention Network proposed by~\citep{liu2021finegrained}, which performs fine-grained fact verification with kernel-based attention. We believe such a graph-based model can better capture information and select a more plausible set of knowledge snippets.

\subsection{Knowledge Grounded Generation}
The third component of our system is to generate responses given select knowledge snippets. Recently, pre-trained language models such as BERT propel progress in natural language generation, but also demonstrate limitation while being directly fine-tuned on small conversation datasets ~\citep{rashkin2019empathetic,wolf2019transfertransfo}. PLATO ~\citep{bao2020plato} was proposed to address this problem by using uni- and bi-directional processing with further pre-training on large-scale Reddit and Twitter conversations. In addition, they introduced a discrete latent variable to grasp one-to-many utterances relationship information between conversations. In DSTC9 track1, ~\citet{tan2020learning} further incorporated knowledge copy method to calculate the probability of response by combining generation distribution with the knowledge-attention distribution.~\citet{tan2020learning} provides an efficient way to generate sentences under the given knowledge, reducing the pressure added on the decoder and is easier for models to generalize to unseen knowledge.

\section{Methodology}
In this section, we first define our problem by dividing it into three separated sub-tasks, then discuss how we address each part of it with different models. Fig \ref{workflow} shows our overall framework.

\subsection{Knowledge-Seeking Turn Detection}
In the first phase of our system, we deploy a binary classifier to decide whether to trigger the knowledge access branch for a given utterance and dialogue history. 

\paragraph{Data Representations} 
To capture whether the current dialog turn will need to trigger the knowledge, we decide to concatenate the current dialog with dialog history. We hope the model can consider richer information than only using one dialog turn. Besides, to make the model understand the speaker of dialog, we added a speaker token ([User] or [Sys]) before every dialog. The speaker tokens represent this dialog turn is spoken by the user or system. We believe that different speakers will provide implicit information to the dialog history and will make our system perform better. Below is the representation of our input data:
\begin{equation}
    [User]U_i[Sys]S_0[User]U_0...[Sys]S_{i-1}
\end{equation}
where $U_i$ equals to the $i_{th}$ turn of user, $S_i$ equals to the $i_{th}$ turn of system.

\paragraph{Binary Classification Model}
In this part, we defined Knowledge-Seeking Turn Detection as a binary classification task. To extract informative features in the dialog context, we chose to use RoBERTa as our encoder since it currently outperformed most pre-trained language models. Besides, we applied a new dialogue turn embedding, which represents the number of turns in the whole dialogue, in our training procedure. We hope the model can learn more from this embedding and regard turn number as important information. After fine-tuning the RoBERTa model, the probability of $x_0$ being the correct answer is calculated as:
\begin{eqnarray}
    e_0 &=& \sum W_h h_0,\\
    p(x_0) &=& \texttt{sigmoid}(e_0),
\end{eqnarray}
where $h_0 \in \mathbb{R}^{d_{h_j}}$ is the output hidden states of [CLS] token, $W_h \in \mathbb{R}^{d_{h_j}}$ are trainable parameters, $p(x_{0})$ is the probability that input dialog will need to trigger the knowledge branch access.

\subsection{Knowledge Selection}
This section describes how we develop our graph-based knowledge selection model GKS. 

\begin{figure*}[t]
\centering
\includegraphics[width=0.6\textwidth]{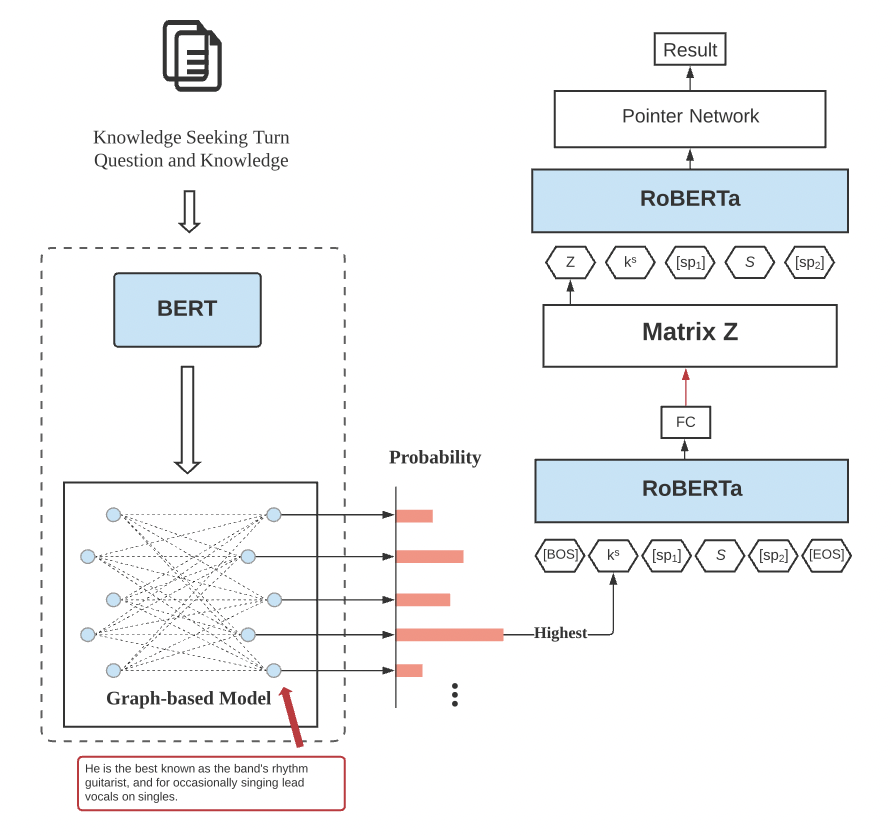} 
\caption{The figure shows the overall framework for knowledge selection and knowledge-grounded generation. The left part describes the workflow of knowledge selection, generating the probability for each node. The generation model takes the highest probability among all nodes as knowledge snippets as input for generation.}
\label{workflow}
\end{figure*}

\paragraph{Knowledge Embedding}
To construct node embedding for GKS, we first build knowledge embedding with the BERT model. For the pre-training BERT model, we first concatenate detected knowledge-needed questions (refer to detected knowledge-seeking turns) and each knowledge with [SEP] token as input, and add [CLS] at the first of every question-knowledge pair. We train the BERT model combined with a linear layer as a binary classification task, referring to  “Related” and “Unrelated.” We then reference this pre-trained BERT model to make node embedding. 
\begin{equation}
    e^n = BERT(k^{n})
\end{equation}
$k^n$ is the $n^{th}$ knowledge in the set $K$ connected with question with [SEP], where $K$ is the knowledge set consists of $k^1...k^n...k^l$ with $l$ knowledge pieces of the current entity. $e^n$ is the hidden state of the $n^{th}$ knowledge-question pair. To elaborate, $e_0^{n}$ represent [CLS]. $e_{1:m}^{n}$ and $e_{1:m+p}^{n}$ represent question and knowledge where question contains $m$ words (tokens) and knowledge $p$ words (tokens).

\paragraph{Graph-Attention Knowledge Selection}
Inspired by \cite{liu2021finegrained}, which successfully captures text information with a graph attention model for factual verification, we develop our selection model with the graph-based model. Based on our aforementioned assumption, we believe the graph-based neural networks can better capture similar traits in knowledge snippets, which resembles clustering for unsupervised learning. The prediction is made per-node, which the node with the highest probability indicates the predicted knowledge. We follow \cite{zhou2019gear}, utilizing node kernel to determine relevance between dialog turn and each knowledge with “readout” function. First, Graph-Attention Knowledge Selector (GKS) applies the translation matrix $M^{q,e^{n}}$ on the $n^{th}$ knowledge hidden state $k^{n}$ and $q$, where $q$ is the question of the user, by the hidden state $e_{1:m}^{n}$ and $e_{1:m+p}^{n}$.
GKS then applies the kernel $Kernel$ match feature on the translation matrix to construct node representation $S(k^n)$ for knowledge selection
\begin{equation}
    S(k^n) = \frac{1}{m}\sum_i^m Kernel(M^{q,k^{n}})
\end{equation}
and
\begin{equation}
    P(k^n|E) = softmax_p (Linear(S(k^n)))
\end{equation}
function as  the readout to calculate $n^{th}$ knowledge selection probability $P(k^n|E)$. The whole model is trained end-to-end by minimizing the cross entropy loss:
\begin{equation}
    L = CrossEntropy(k^*, P(k^p|E))
\end{equation}
where $k^*$ is the golden knowledge to the detect knowledge turn.

\begin{table} 
\begin{adjustbox}{width={\columnwidth},center}%
\centering
\small
\begin{tabular}{lccccc}
\hline
\textbf{Model}            & \textbf{Recall}   & \textbf {Precision}    & \textbf{F1} \\ \hline
Baseline                        & 0.9992          & 0.9719
&0.9853 \\
ROBERTA-HS                     & 0.9981         & 0.9963          &    0.9972                                \\
ROBERTA-WD (DA)                 & \textbf{0.9996}          & \textbf{0.9985}          & \textbf{0.9990}                     \\
\cite{he2021learning}           & 0.99102        & 0.9969            & 0.9939                              \\
\cite{Tang:21}                   & 0.9817         & 0.9465              & 0.9638              \\ \hline
Ours & 0.9918          & 0.9921 & 0.9920                                \\ \hline

\end{tabular}
\end{adjustbox}
\caption{\label{auto2}
Our result of Knowledge-Seeking Turn Detection.
 Where ROBERTA-HS and ROBERTA-WD (DA) are from \citet{Mi:21}.
}
\label{detect9}
\end{table}

\subsection{Knowledge Grounded Generation}
After candidate knowledge snippets are given, we then select the one snippet with the highest probability as an input for knowledge grounded generation. Inspired by ~\citet{bao2020plato} and ~\citet{tan2020learning}, we then leverage RoBERTa-based architecture with the consideration of latent variable. Inspired by ~\citet{tan2020learning}, we concatenate knowledge snippet and dialogue history with special tokens as input. Unlike ~\citet{tan2020learning}, we consider both questions and answer part of the knowledge snippet:

\begin{equation}
\begin{adjustbox}{width={\columnwidth}}
$<BOS>k^s<sp1>s_{1}<sp2>s_{2}...<sp2>r<EOS>$
\end{adjustbox}
\end{equation}

where $k^s$ represents selected knowledge snippet, $<sp_{1}>$ and $<sp_{2}>$ represent two speakers in the dialogue respectively, and $s_{n}$ denotes $n^{th}$ term in dialog history . $r$ represents the response.
 Following ~\citet{tan2020learning}, we encode response into the Z matrix, of which each row represents a special z corresponding to given examples. To select a specific z as our latent variable, we estimate posterior probability $q_{\phi} (z|S, k^{s}, r)$, where S denotes dialogue history. The rest of the architecture and calculation process, such as knowledge copy mechanism, segmented response generation, and modified beam search, are essentially identical to the ones in ~\citet{tan2020learning}. We illustrate it with the subtask2 model in Figure \ref{workflow}.

\section{Experiments}
This section demonstrates our experiment results. For the Baseline and chosen baselines, we report from their paper presented in DSTC9 last year. 

\subsection{Knowledge Seeking-Turn Detection}
Table \ref{detect9} shows our result on the DSTC9 data-set. As the proposed baseline model of DSTC9 is already performing very well, other proposed models have only a slight difference in their results. Our experiment results show that our model outperformed several baselines on the F1 score, which indicates that our approach on the training model with [User] and [Sys] tokens gives the language model more ability to learn user utterances patterns. ~\citet{Mi:21} perform better from selected models. We assume their proposed training strategy with data augmentation is the key reason to gain performance under the condition that most language models could gain excellent performance on the original data-set.

\begin{table} 
\begin{adjustbox}{width={\columnwidth},center}%
\centering
\small
\begin{tabular}{lccccc}
\hline
\textbf{Model}            & \textbf{Acc@5}   & \textbf{Acc@1}    & \textbf{MRR@5}\\ \hline
Baseline                        & 0.8772          & 0.6201
&0.7263 \\ROBERTA-WD                     & 0.9745          & 0.9145          &    0.9428    \\ROBERTA-WD (IS)     & 0.9741          & 0.9456           & 0.9589  \\ROBERTA-WD-listwise   & 0.9752    & 0.9394  & 0.9566             \\
\cite{he2021learning}                & 0.9892 & 0.9465                     & \textbf{0.9665} 
    \\
\cite{Tang:21}                   & 0.9665         & 0.9117              & 0.9372              \\ \hline
KGS & \textbf{0.9899}          & \textbf{0.95435} &  -                              \\ \hline
\end{tabular}
\end{adjustbox}
\caption{Result of our presented Knowledge Selection model KGS. ROBERTA-WD, ROBERTA-WD (IS), and ROBERTA-WD-listwise are proposed in \citet{Mi:21}.}
\label{selection9}
\end{table}

\subsection{Knowledge Selection} 
Since our motivation is aiming to develop a better solution for the knowledge section without noise in spoken language translation and preventing potential defects mentioned in earlier sections that previous approaches don’t cope with, we further test our proposed model on the DSTC9 dataset. Table \ref{selection9} shows the final result of KGS model performance on the DSTC9 track one dataset. We selected several models proposed in last year's competition and SOTA models as the baseline. ROBERTA-WD (IS)  in \cite{Mi:21} used sampling technique and k-fold cross-validation during the training process. \citep{he2021learning} acquired multi-scale negatives to replace random sampling, which might lead to coarse-grained class separation. \cite{Tang:21} is an ELECTRA-based model with proposed aggregated loss, which contains the correlation between the domains, entity names, knowledge snippets, and dialog contexts. The result shows that our model, which applies a graph-based model in the selection process, outperforms past approaches that only rely on language models, even without data augmentation. The results suggested that our proposed graph-based architecture did enhance the performance as our settings on knowledge embedding generation was simpler than other SOTA models.

\begin{table*} 
\centering
\begin{tabular}{ c c c c c c c c c c }
\textbf{Model} & \textbf{BLEU-1} & \textbf{BLEU-2} & \textbf{BLEU-3} & \textbf{BLEU-4} & \textbf{METEOR} & \textbf{ROUGE-1} & \textbf{ROUGE-2} & \textbf{ROUGE-L} \\
\hline
Baseline & 0.3601&0.2202&0.1389&0.0956&0.3600&0.3939&0.1749&0.3501\\
\cite{Mi:21} & 0.4330 & \textbf{0.3061} & \textbf{0.2133}  & \textbf{0.1616} & \textbf{0.4535} & \textbf{0.4795} & \textbf{0.2520} & \textbf{0.4304} \\
\cite{Tang:21} & 0.3684 & 0.2374 & 0.1531  & 0.1030 & 0.3719 & 0.4113 & 0.1938 &0.3692\\
\cite{he2021learning} &0.4267 &0.2789 &0.1858 &0.1357 & 0.4324&0.4587 &0.2249 & 0.4093 \\
\hline
Ours & \textbf{0.4356} & 0.2978 &  0.1993  &  0.1378 & 0.4400 & 0.4711 & 0.2415 &  0.4262  \\

\end{tabular}
\caption{\label{auto4}
Automatic metric of our generation model against other baselines in subtask3.
}
\label{generation9}
\end{table*}

\subsection{Knowledge Grounded Generation}
The generated results are demonstrated in Table \ref{generation9}, it is commensurate with others in DSTC9.
Following \citet{tan2020learning}, our RoBERTa-based model has the same hyperparameters as the baseline model in \citet{kim2021domain}. The learning rate is $6.25e-5$, the batch size is 4, and the number of gradient accumulation steps is 32. The number of hidden variable z is set to 5. Our model is trained in 20 epochs, and we use a copy mechanism followed by vanilla beam search to get our final generated result.


\section{Conclusions}

In this paper, we proposed a framework for DSTC10 and DSTC9. Our main goal is to develop a better solution for knowledge selection tasks, which only rely on language models to perform selection in the past. The results showed that our proposed knowledge selection model with a graph-based model performed better than the 
proposed models last year. For our future goal, we are interested in replacing knowledge turn question embedding, which is constructed with text sentence embedding in our original setting, with wave embedding. We assume this could better obtain spoken without hurting the overall system.

\bibliography{aaai22}

\begin{thebibliography}{23}
\providecommand{\natexlab}[1]{#1}

\bibitem[{Bao et~al.(2020)Bao, He, Wang, Wu, and Wang}]{bao2020plato}
Bao, S.; He, H.; Wang, F.; Wu, H.; and Wang, H. 2020.
\newblock PLATO: Pre-trained Dialogue Generation Model with Discrete Latent
  Variable.
\newblock arXiv:1910.07931.

\bibitem[{Budzianowski and Vulic(2019)}]{Budzianowski:19}
Budzianowski, P.; and Vulic, I. 2019.
\newblock Hello, It's {GPT-2} - How Can {I} Help You? Towards the Use of
  Pretrained Language Models for Task-Oriented Dialogue Systems.
\newblock \emph{CoRR}, abs/1907.05774.

\bibitem[{Clark et~al.(2020)Clark, Luong, Le, and Manning}]{clark2020electra}
Clark, K.; Luong, M.-T.; Le, Q.~V.; and Manning, C.~D. 2020.
\newblock ELECTRA: Pre-training Text Encoders as Discriminators Rather Than
  Generators.
\newblock arXiv:2003.10555.

\bibitem[{Devlin et~al.(2019)Devlin, Chang, Lee, and Toutanova}]{Devlin:19}
Devlin, J.; Chang, M.-W.; Lee, K.; and Toutanova, K. 2019.
\newblock {BERT}: Pre-training of Deep Bidirectional Transformers for Language
  Understanding.
\newblock In \emph{Proceedings of the 2019 Conference of the North {A}merican
  Chapter of the Association for Computational Linguistics: Human Language
  Technologies, Volume 1 (Long and Short Papers)}, 4171--4186. Minneapolis,
  Minnesota: Association for Computational Linguistics.

\bibitem[{Dong et~al.(2019)Dong, Yang, Wang, Wei, Liu, Wang, Gao, Zhou, and
  Hon}]{dong2019unified}
Dong, L.; Yang, N.; Wang, W.; Wei, F.; Liu, X.; Wang, Y.; Gao, J.; Zhou, M.;
  and Hon, H.-W. 2019.
\newblock Unified Language Model Pre-training for Natural Language
  Understanding and Generation.
\newblock arXiv:1905.03197.

\bibitem[{Eric et~al.(2019)Eric, Goel, Paul, Kumar, Sethi, Ku, Goyal, Agarwal,
  Gao, and Hakkani-Tur}]{eric2019multiwoz}
Eric, M.; Goel, R.; Paul, S.; Kumar, A.; Sethi, A.; Ku, P.; Goyal, A.~K.;
  Agarwal, S.; Gao, S.; and Hakkani-Tur, D. 2019.
\newblock MultiWOZ 2.1: A Consolidated Multi-Domain Dialogue Dataset with State
  Corrections and State Tracking Baselines.
\newblock arXiv:1907.01669.

\bibitem[{Ham et~al.(2020)Ham, Lee, Jang, and Kim}]{Ham:20}
Ham, D.; Lee, J.-G.; Jang, Y.; and Kim, K.-E. 2020.
\newblock End-to-End Neural Pipeline for Goal-Oriented Dialogue Systems using
  {GPT}-2.
\newblock In \emph{Proceedings of the 58th Annual Meeting of the Association
  for Computational Linguistics}, 583--592. Online: Association for
  Computational Linguistics.

\bibitem[{He et~al.(2021)He, Lu, Bao, Wang, Wu, Niu, and Wang}]{he2021learning}
He, H.; Lu, H.; Bao, S.; Wang, F.; Wu, H.; Niu, Z.; and Wang, H. 2021.
\newblock Learning to Select External Knowledge with Multi-Scale Negative
  Sampling.
\newblock arXiv:2102.02096.

\bibitem[{Kim et~al.(2020)Kim, Eric, Gopalakrishnan, Hedayatnia, Liu, and
  Hakkani-Tur}]{kim2020domain}
Kim, S.; Eric, M.; Gopalakrishnan, K.; Hedayatnia, B.; Liu, Y.; and
  Hakkani-Tur, D. 2020.
\newblock Beyond Domain APIs: Task-oriented Conversational Modeling with
  Unstructured Knowledge Access.
\newblock arXiv:2006.03533.

\bibitem[{Kim et~al.(2021)Kim, Eric, Hedayatnia, Gopalakrishnan, Liu, Huang,
  and Hakkani-Tur}]{kim2021domain}
Kim, S.; Eric, M.; Hedayatnia, B.; Gopalakrishnan, K.; Liu, Y.; Huang, C.-W.;
  and Hakkani-Tur, D. 2021.
\newblock Beyond Domain APIs: Task-oriented Conversational Modeling with
  Unstructured Knowledge Access Track in DSTC9.
\newblock arXiv:2101.09276.

\bibitem[{Liu et~al.(2019)Liu, Ott, Goyal, Du, Joshi, Chen, Levy, Lewis,
  Zettlemoyer, and Stoyanov}]{Yinhan:19}
Liu, Y.; Ott, M.; Goyal, N.; Du, J.; Joshi, M.; Chen, D.; Levy, O.; Lewis, M.;
  Zettlemoyer, L.; and Stoyanov, V. 2019.
\newblock RoBERTa: {A} Robustly Optimized {BERT} Pretraining Approach.
\newblock \emph{CoRR}, abs/1907.11692.

\bibitem[{Liu et~al.(2021)Liu, Xiong, Sun, and Liu}]{liu2021finegrained}
Liu, Z.; Xiong, C.; Sun, M.; and Liu, Z. 2021.
\newblock Fine-grained Fact Verification with Kernel Graph Attention Network.
\newblock arXiv:1910.09796.

\bibitem[{Mi et~al.(2021)Mi, Ren, Dai, He, Sun, Li, Zheng, and Xu}]{Mi:21}
Mi, H.; Ren, Q.; Dai, Y.; He, Y.; Sun, J.; Li, Y.; Zheng, J.; and Xu, P. 2021.
\newblock Towards Generalized Models for Beyond Domain API Task-oriented
  Dialogue.
\newblock \emph{DSTC 9}.

\bibitem[{Radford et~al.(2018)Radford, Wu, Child, Luan, Amodei, and
  Sutskever}]{Radford:18}
Radford, A.; Wu, J.; Child, R.; Luan, D.; Amodei, D.; and Sutskever, I. 2018.
\newblock Language Models are Unsupervised Multitask Learners.
\newblock \emph{OpenAI blog}.

\bibitem[{Ramos(2003)}]{Ramos2003UsingTT}
Ramos, J.~E. 2003.
\newblock Using TF-IDF to Determine Word Relevance in Document Queries.
\newblock In \emph{Proceedings of the first instructional conference on machine
  learning}.

\bibitem[{Rashkin et~al.(2019)Rashkin, Smith, Li, and
  Boureau}]{rashkin2019empathetic}
Rashkin, H.; Smith, E.~M.; Li, M.; and Boureau, Y.-L. 2019.
\newblock Towards Empathetic Open-domain Conversation Models: a New Benchmark
  and Dataset.
\newblock arXiv:1811.00207.

\bibitem[{Rastogi et~al.(2020)Rastogi, Zang, Sunkara, Gupta, and
  Khaitan}]{rastogi2020schemaguided}
Rastogi, A.; Zang, X.; Sunkara, S.; Gupta, R.; and Khaitan, P. 2020.
\newblock Schema-Guided Dialogue State Tracking Task at DSTC8.
\newblock arXiv:2002.01359.

\bibitem[{Shah et~al.(2019)Shah, Gupta, Fayazi, and
  Hakkani-Tur}]{shah2019robust}
Shah, D.~J.; Gupta, R.; Fayazi, A.~A.; and Hakkani-Tur, D. 2019.
\newblock Robust Zero-Shot Cross-Domain Slot Filling with Example Values.
\newblock arXiv:1906.06870.

\bibitem[{Tan et~al.(2020)Tan, Yang, Zheng, Li, Feng, Gu, Liu, Liu, Ling, and
  Zhu}]{tan2020learning}
Tan, C.-H.; Yang, X.; Zheng, Z.; Li, T.; Feng, Y.; Gu, J.-C.; Liu, Q.; Liu, D.;
  Ling, Z.-H.; and Zhu, X. 2020.
\newblock Learning to Retrieve Entity-Aware Knowledge and Generate Responses
  with Copy Mechanism for Task-Oriented Dialogue Systems.
\newblock arXiv:2012.11937.

\bibitem[{Tang et~al.(2021)Tang, Shang, Lv, Fu, Zhang, Huang, and
  Zhang}]{Tang:21}
Tang, L.; Shang, Q.; Lv, K.; Fu, Z.; Zhang, S.; Huang, C.; and Zhang, Z. 2021.
\newblock RADGE: Relevance Learning and Generation Evaluating Method for
  Task-Oriented Conversational Systems.
\newblock \emph{DSTC 9}.

\bibitem[{Wang et~al.(2018)Wang, Singh, Michael, Hill, Levy, and
  Bowman}]{Wang:18}
Wang, A.; Singh, A.; Michael, J.; Hill, F.; Levy, O.; and Bowman, S.~R. 2018.
\newblock {GLUE:} {A} Multi-Task Benchmark and Analysis Platform for Natural
  Language Understanding.
\newblock \emph{CoRR}, abs/1804.07461.

\bibitem[{Wolf et~al.(2019)Wolf, Sanh, Chaumond, and
  Delangue}]{wolf2019transfertransfo}
Wolf, T.; Sanh, V.; Chaumond, J.; and Delangue, C. 2019.
\newblock TransferTransfo: A Transfer Learning Approach for Neural Network
  Based Conversational Agents.
\newblock arXiv:1901.08149.

\bibitem[{Zhou et~al.(2019)Zhou, Han, Yang, Liu, Wang, Li, and
  Sun}]{zhou2019gear}
Zhou, J.; Han, X.; Yang, C.; Liu, Z.; Wang, L.; Li, C.; and Sun, M. 2019.
\newblock GEAR: Graph-based Evidence Aggregating and Reasoning for Fact
  Verification.
\newblock arXiv:1908.01843.

\end{thebibliography}

\end{document}